%% file: ms.tex
\documentclass[10pt,twocolumn,letterpaper]{article}

\usepackage{cvpr}
\usepackage{times}
\usepackage{epsfig}
\usepackage{graphicx}
\usepackage{amsmath}
\usepackage{amssymb}
\usepackage{paralist} % for compactitem
\usepackage{multirow}
\usepackage[pagebackref=true,breaklinks=true,letterpaper=true,colorlinks,bookmarks=false]{hyperref}

\cvprfinalcopy % *** Uncomment this line for the final submission

\def\cvprPaperID{1063} % *** Enter the CVPR Paper ID here

\ifcvprfinal\fi
\begin{document}

%%%%%%%%% TITLE
\title{AVA: A Video Dataset of Spatio-temporally Localized Atomic Visual Actions}

\author{Chunhui Gu\thanks{Google Research} \and Chen Sun\footnotemark[1] \and David A. Ross\footnotemark[1] \and Carl Vondrick\footnotemark[1] \and Caroline Pantofaru\footnotemark[1] \and  Yeqing Li\footnotemark[1] \and Sudheendra Vijayanarasimhan\footnotemark[1] \and George Toderici\footnotemark[1] \and Susanna Ricco\footnotemark[1] \and Rahul Sukthankar\footnotemark[1] \and Cordelia Schmid\thanks{Inria, Laboratoire Jean Kuntzmann, Grenoble, France}\textsuperscript{ } \footnotemark[1] \and Jitendra Malik\thanks{University of California at Berkeley, USA}\textsuperscript{ } \footnotemark[1]}

\maketitle
\thispagestyle{empty}

% Define numbers as variables that are referenced multiple times here.
\newcommand{\numClasses}{80}
\newcommand{\numMovies}{430}
\newcommand{\numClips}{386k}
\newcommand{\numBBoxes}{614k}
\newcommand{\numActions}{1.58M}

% Comment out second line to disable todos.
\newcommand{\todo}[1]{}
\renewcommand{\todo}[1]{{\color{red} TODO: {#1}}}

\begin{abstract}
  \input{abstract.tex}
\end{abstract}

\input{introduction}

\input{related}

\input{collection}

\input{dataset}

\input{experiments}

\input{conclusion}

\vspace{1mm}

\noindent\textbf{Acknowledgement} We thank Abhinav Gupta, Abhinav Shrivastava, Andrew Gallagher, Irfan Essa, and Vicky Kalogeiton for discussion and comments about this work.

{\small
\bibliographystyle{ieee}
\bibliography{egbib}
}

\newpage
\input{supplemental}

\end{document}

%% file: abstract.tex
This paper introduces a video dataset of spatio-temporally localized Atomic Visual Actions (AVA). The AVA dataset densely annotates \numClasses{} atomic visual actions in \numMovies{} 15-minute video clips, where actions are localized in space and time, resulting in \numActions{} action labels with multiple labels per person occurring frequently. The key characteristics of our dataset are: (1) the definition of atomic visual actions,  rather than composite actions; (2) precise spatio-temporal annotations with possibly multiple annotations for each person; (3) exhaustive annotation of these atomic actions over 15-minute video clips; (4) people temporally linked across consecutive segments; and (5) using movies to gather a varied set of action representations. This departs from existing datasets for spatio-temporal action recognition, which typically provide sparse annotations for composite actions in short video clips.

AVA, with its realistic scene and action complexity, exposes the intrinsic difficulty of action recognition. To benchmark this, we present a novel approach for action localization that builds upon the current state-of-the-art methods, and demonstrates better performance on JHMDB and UCF101-24 categories. While setting a new state of the art on existing datasets, the overall results on AVA are low at 15.6\% mAP, underscoring the need for developing new approaches for video understanding.

%% file: introduction.tex
\vspace{-2em}
\section{Introduction}
\label{sec:introduction}

\definecolor{orange}{rgb}{1, 0.5, 0}

\begin{figure}[t]
\includegraphics[width=.95\columnwidth]{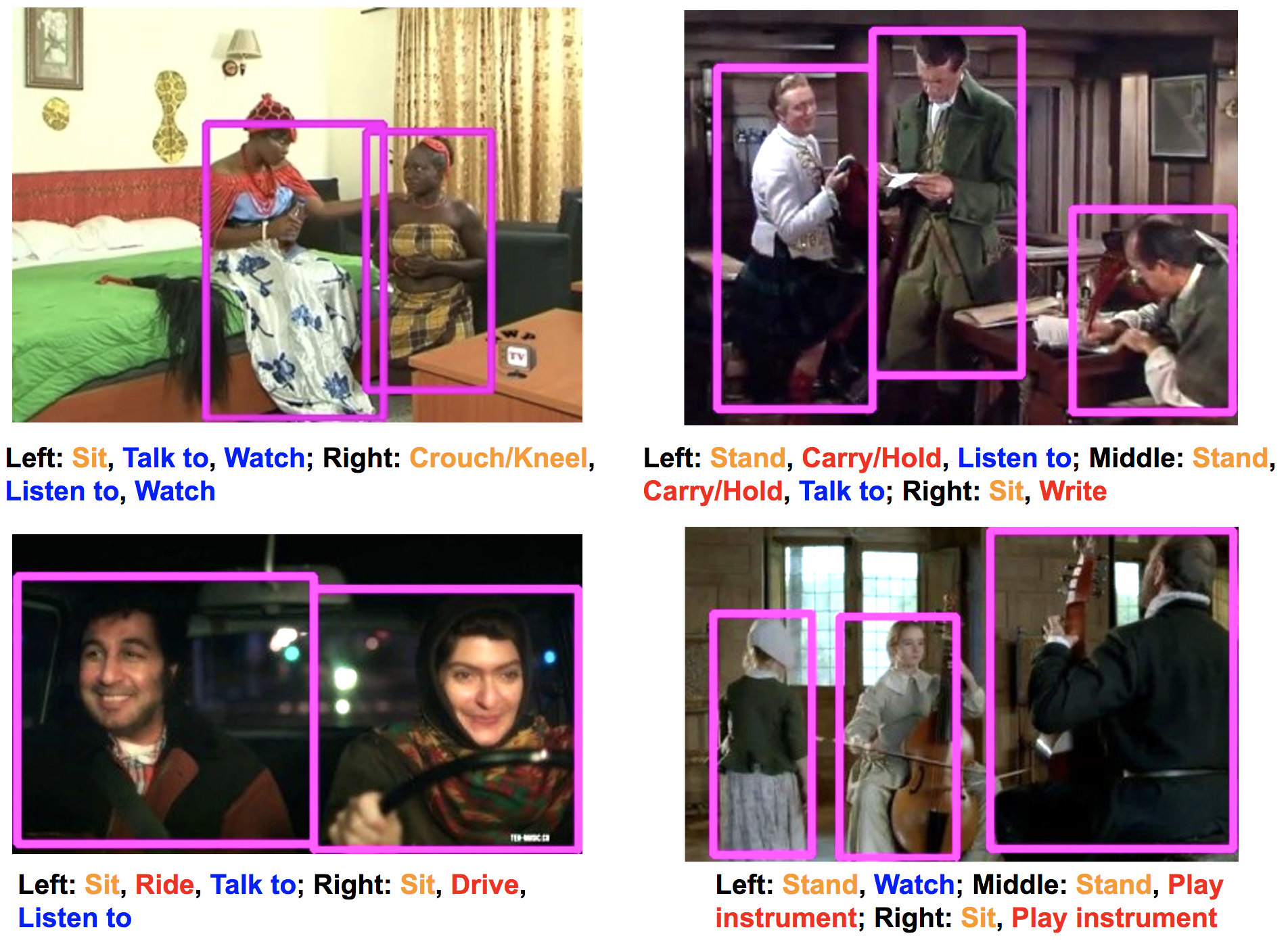}
\caption{The bounding box and action annotations in sample frames of the AVA dataset. Each bounding box is associated with 1 pose action (in \textcolor{orange}{orange}), 0--3 interactions with objects (in \textcolor{red}{red}), and 0--3 interactions with other people (in \textcolor{blue}{blue}). Note that some of these actions require temporal context to accurately label.
\label{fig:teaser}
}
\vspace{-1em}
\end{figure}

We introduce a new annotated video dataset, AVA, to advance action recognition research (see Fig.~\ref{fig:teaser}). The annotation is person-centric at a sampling frequency of 1~Hz. Every person is localized using a bounding box and the attached labels correspond to (possibly multiple) actions being performed by the actor: one action corresponding to the actor's \textbf{\textcolor{orange}{pose}} (orange text) --- standing, sitting, walking, swimming etc.\ --- and there may be additional actions corresponding to \textbf{\textcolor{red}{interactions with objects}} (red text) or \textbf{\textcolor{blue}{interactions with other persons}} (blue text). Each person in a frame containing multiple actors is labeled separately.

To label the actions performed by a person, a key choice is the annotation vocabulary, which in turn is determined by the temporal granularity at which actions are classified. We use short segments ($\pm 1.5$ seconds centered on a keyframe) to provide temporal context for labeling the actions in the middle frame. This enables the annotator to use movement cues for disambiguating actions such as pick up or put down that cannot be resolved in a static frame. We keep the temporal context relatively brief because we are interested in (temporally) fine-scale  annotation of physical actions, which  motivates ``Atomic Visual Actions'' (AVA). The vocabulary consists of \numClasses{} different atomic visual actions. Our dataset is sourced from the 15th to 30th minute time intervals of \numMovies{} different movies, which given the 1~Hz sampling frequency gives us nearly 900 keyframes for each movie.  In each keyframe, every person is labeled with (possibly multiple) actions from the AVA vocabulary. Each person is linked to the consecutive keyframes to provide short temporal sequences of action labels (Section~\ref{sec:temp_structure}). We now motivate the main design choices of AVA.

\begin{figure}[t]
\vspace{-0.5em}
\centerline{\includegraphics[width=.95\columnwidth]{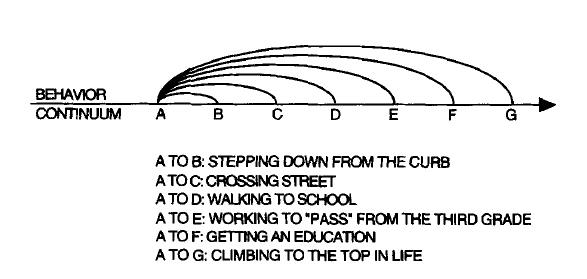}}
\vspace{-0.5em}
\caption{This figure illustrates the hierarchical nature of an activity. From Barker and Wright~\cite{Barker1954}, pg.~247.}
\label{fig:barker-wright}
\vspace{-1em}
\end{figure}

\noindent \textbf{Atomic action categories.} Barker \& Wright~\cite{Barker1954} noted the hierarchical nature of activity (Fig.~\ref{fig:barker-wright}) in their classic study of the "behavior episodes" in the daily lives of the residents of a small town in Kansas.  At the finest level, the actions consist of atomic body movements or object manipulation but at coarser levels, the most natural descriptions are in terms of intentionality and goal-directed behavior. 

This hierarchy makes defining a vocabulary of action labels ill posed, contributing to the slower progress of our field compared to object recognition; exhaustively listing high-level behavioral episodes is impractical. However if we limit ourselves to fine time scales, then the actions are very physical in nature and have clear visual signatures. Here, we annotate keyframes at 1~Hz as this is sufficiently dense to capture the complete semantic content of actions while enabling us to avoid requiring unrealistically precise temporal annotation of action boundaries. The THUMOS challenge~\cite{THUMOS} observed that action boundaries (unlike objects) are inherently fuzzy, leading to significant inter-annotator disagreement. By contrast, annotators can easily determine (using $\pm$1.5s of context) whether a frame \emph{contains} a given action. Effectively, AVA localizes action start and end points to an acceptable precision of $\pm 0.5~s$.

\noindent \textbf{Person-centric action time series.} While events such as trees falling do not involve people, our focus is on the activities of people, treated as single agents. There could be multiple people as in sports or two people hugging, but each one is an agent with individual choices, so we treat each separately. The action labels assigned to a person over time is a rich source of data for temporal modeling (Section~\ref{sec:temp_structure}).

\noindent \textbf{Annotation of movies.} Ideally we would want behavior ``in the wild''. We do not have that, but movies are a compelling approximation, particularly when we consider the diversity of genres and countries with flourishing film industries. We do expect some bias in this process. Stories have to be interesting and there is a grammar of the film language~\cite{Arijon1991} that communicates through the juxtaposition of shots. That said, in each shot we can expect an unfolding sequence of human actions, somewhat representative of reality, as conveyed by competent actors. AVA complements the current datasets sourced from user generated video because we expect movies to contain a greater range of activities as befits the telling of diverse stories.

\noindent \textbf{Exhaustive action labeling.}
We label all the actions of all the people in all the keyframes. This will naturally result in a Zipf's law type of imbalance across action categories. There will be many more examples of typical actions (standing or sitting) than memorable ones (dancing), but this is how it should be!  Recognition models need to operate on realistic ``long tailed'' action distributions~\cite{long_tail17} rather than being scaffolded using artificially balanced datasets. Another consequence of our protocol is that since we do not retrieve examples of action categories by explicit querying of internet video resources, we avoid a certain kind of bias: opening a door is a common event that occurs frequently in movie clips; however a door opening action that has been tagged as such on YouTube is likely attention worthy in a way that makes it atypical.

We believe that AVA, with its realistic complexity, exposes the inherent difficulty of action recognition hidden by many popular datasets in the field. A video clip  of a single person performing a visually salient action like swimming in typical background is  easy to discriminate from, say, one of a person running. Compare with AVA where we encounter multiple actors, small in image size, performing actions which are only subtly different such as touching vs.\ holding an object. To verify this intuition, we do comparative bench-marking on JHMDB~\cite{jhmdb}, UCF101-24 categories~\cite{ucfsports} and AVA. The approach we use for spatio-temporal action localization (see Section~\ref{sec:I3D}) builds upon multi-frame approaches~\cite{T_CNN_iccv17,tubelets_iccv17}, but classifies tubelets with I3D convolutions~\cite{i3d_cvpr17}. We obtain state-of-the-art performance on JHMDB~\cite{jhmdb} and UCF101-24 categories~\cite{ucfsports} (see Section~\ref{sec:experiments}) while the mAP on AVA is only 15.6\%. 

The AVA dataset has been released publicly at \url{https://research.google.com/ava/}.
\vspace{-0.8em}

%% file: related.tex
\section{Related work}
\label{sec:related}

\noindent{\bf Action recognition datasets.} 
Most popular action classification datasets, such as KTH~\cite{Schuldt2004}, Weizmann~\cite{Blank2005}, Hollywood-2~\cite{Marszalek2009}, HMDB~\cite{kuehne2011}, UCF101~\cite{ucf101} consist of short clips, manually trimmed to capture a single action. These datasets are ideally suited for training fully-supervised, whole-clip, forced-choice video classifiers. Recently, datasets, such as TrecVid MED~\cite{2014trecvid},  Sports-1M~\cite{Karpathy2014}, YouTube-8M~\cite{youtube8m}, Something-something~\cite{something_iccv17}, SLAC \cite{zhao2017slac}, Moments in Time \cite{monfortmoments}, and Kinetics~\cite{kinetics17} have focused on large-scale video classification, often with automatically generated -- and hence potentially noisy -- annotations. They serve a valuable purpose but address a different need than AVA. 

Some recent work has moved towards temporal localization. ActivityNet~\cite{activitynet}, THUMOS~\cite{THUMOS}, MultiTHUMOS~\cite{MultiTHUMOS} and Charades~\cite{charades2016} use large numbers of untrimmed videos, each containing multiple actions, obtained either from YouTube (ActivityNet, THUMOS, MultiTHUMOS) or from crowdsourced actors (Charades). The datasets provide temporal (but not spatial) localization for each action of interest.  AVA differs from them, as we provide spatio-temporal annotations for each subject performing an action and annotations are dense over 15-minute clips. 

A few datasets, such as CMU~\cite{Ke2005}, MSR Actions~\cite{Yuan2009}, UCF Sports~\cite{ucfsports} and JHMDB~\cite{jhmdb} provide spatio-temporal annotations in each frame for short videos. The main differences with our AVA dataset are: the small number of actions; the small number of video clips; and the fact that clips are very short. Furthermore, actions are composite (e.g., pole-vaulting) and not atomic as in AVA. Recent extensions, such as UCF101~\cite{ucf101}, DALY~\cite{weinzaepfel2016} and Hollywood2Tubes~\cite{mettes2016} evaluate spatio-temporal localization in untrimmed videos, which makes the task significantly harder and results in a performance drop. However, the action vocabulary is still restricted to a limited number of composite actions. Moreover, they do not densely cover the actions; a good example is BasketballDunk in UCF101, where only the dunking player is annotated. However, real-world applications often require a continuous annotations of atomic actions of all humans, which can then be composed into higher-level events. This motivates AVA's exhaustive labeling over 15-minute clips.

AVA is also related to still image action recognition datasets~\cite{hico2015,PASCAL,gupta2015} that are limited in two ways. First, the lack of motion can make action disambiguation difficult. Second, modeling composite events as a \emph{sequence} of atomic actions is not possible in still images. This is arguably out of scope here, but clearly required in many real-world applications, for which AVA does provide training data.

\noindent{\bf Methods for spatio-temporal action localization.}
Most recent approaches~\cite{gkioxari2015,peng2016multi,saha2016,weinzaepfel2015} rely on object detectors trained to discriminate action classes at the frame level with a two-stream variant, processing RGB and flow data separately. The resulting per-frame detections are then linked using dynamic programming~\cite{gkioxari2015,Singh_ICCV2017} or tracking~\cite{weinzaepfel2015}. All these approaches rely on integrating frame-level detections. Very recently, multi-frame approaches have emerged: Tubelets~\cite{tubelets_iccv17} jointly estimate localization and classification over several frames, T-CNN~\cite{T_CNN_iccv17} use 3D convolutions to estimate short tubes, micro-tubes rely on two successive frames~\cite{micro_tube2017} and pose-guided 3D convolutions add pose to a two-stream approach~\cite{pose_brox2017}. We build upon the idea of spatio-temporal tubes, but employ state-of-the-art I3D convolution~\cite{i3d_cvpr17} and Faster R-CNN~\cite{ren2015faster} region proposals to outperform the state of the art.
\vspace{-0.8em}

%% file: collection.tex
\section{Data collection}
\label{sec:collection}

\begin{figure}[t]
\centerline{\includegraphics[width=.9\columnwidth]{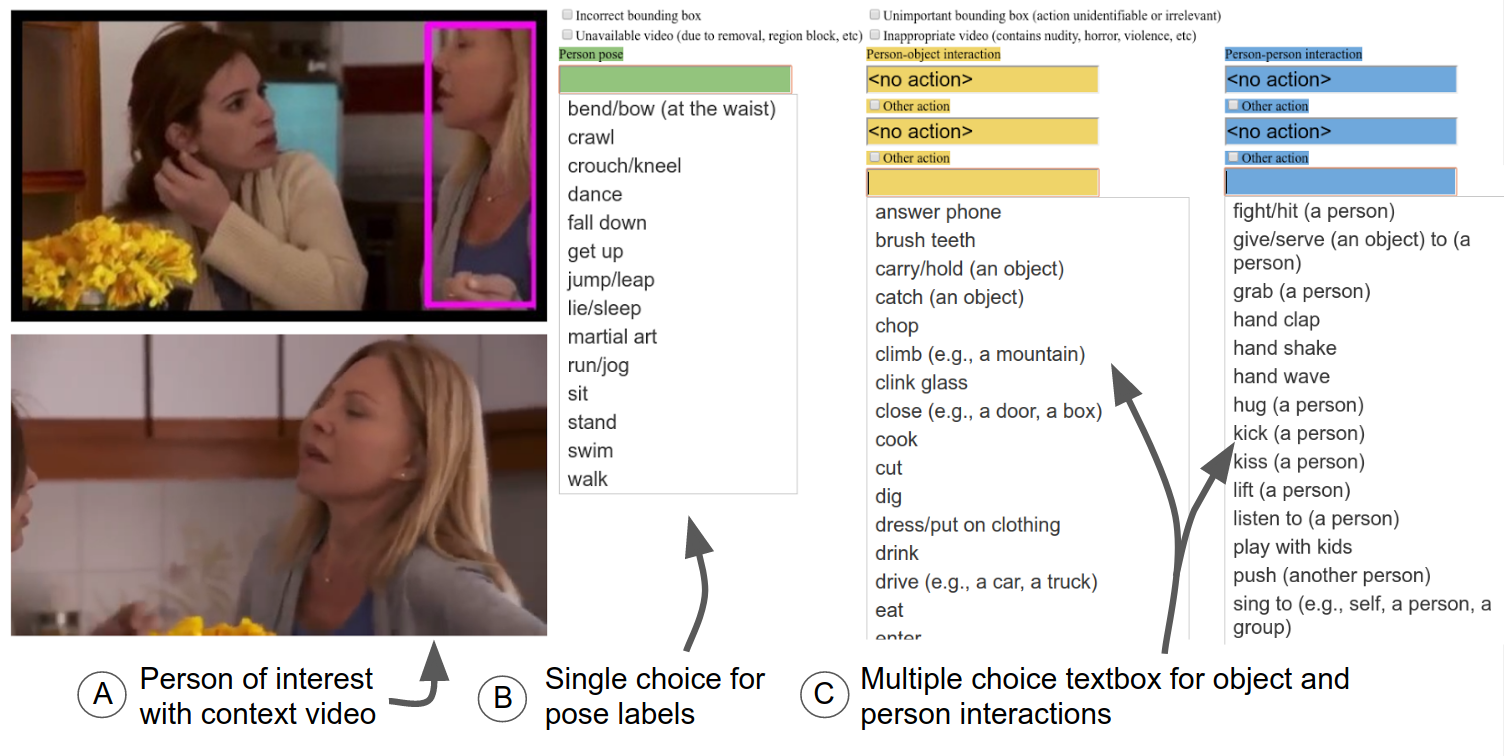}}
\caption{User interface for action annotation. Details in Sec~\ref{sec:action_label}.}
\label{fig:annotation}
\vspace{-1em}
\end{figure}

Annotation of the AVA dataset consists of five stages: action vocabulary generation, movie and segment selection, person bounding box annotation, person linking and action annotation.

\subsection{Action vocabulary generation}

We follow three principles to generate our action vocabulary. The first one is generality. We collect generic actions in daily-life scenes, as opposed to specific activities in specific environments (e.g., playing basketball on a basketball court). The second one is atomicity. Our action classes have clear visual signatures, and are typically independent of interacted objects (e.g., hold without specifying what object to hold). This keeps our list short yet complete. The last one is exhaustivity. We initialized our list using knowledge from previous datasets, and iterated the list in several rounds until it covered $\sim$99\% of actions in the AVA dataset labeled by annotators. We end up with 14 pose classes, 49 person-object interaction classes and 17 person-person interaction classes in the vocabulary.

\begin{figure*}[tb!]
    \centering
    \small
\begin{minipage}{0.32\linewidth}\centering\includegraphics[width=\columnwidth]{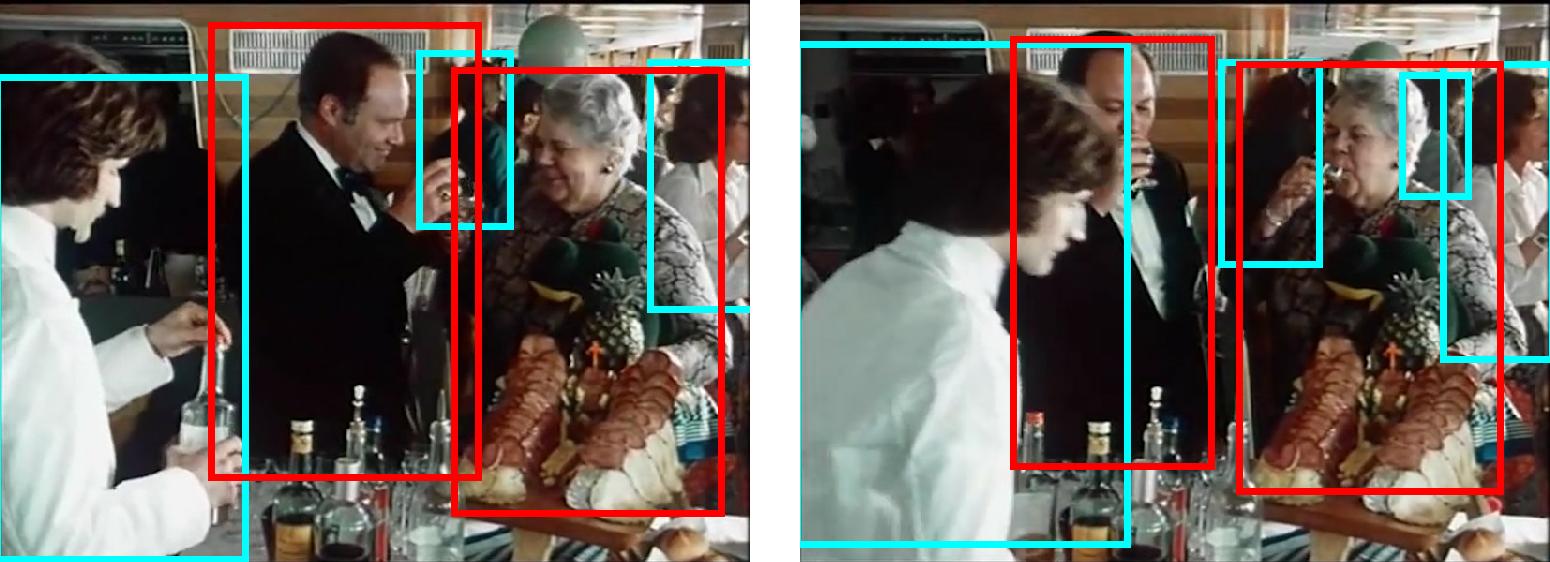} \\[-0.12cm]
clink glass $\rightarrow$ drink \\[0.11cm]
\includegraphics[width=\columnwidth]{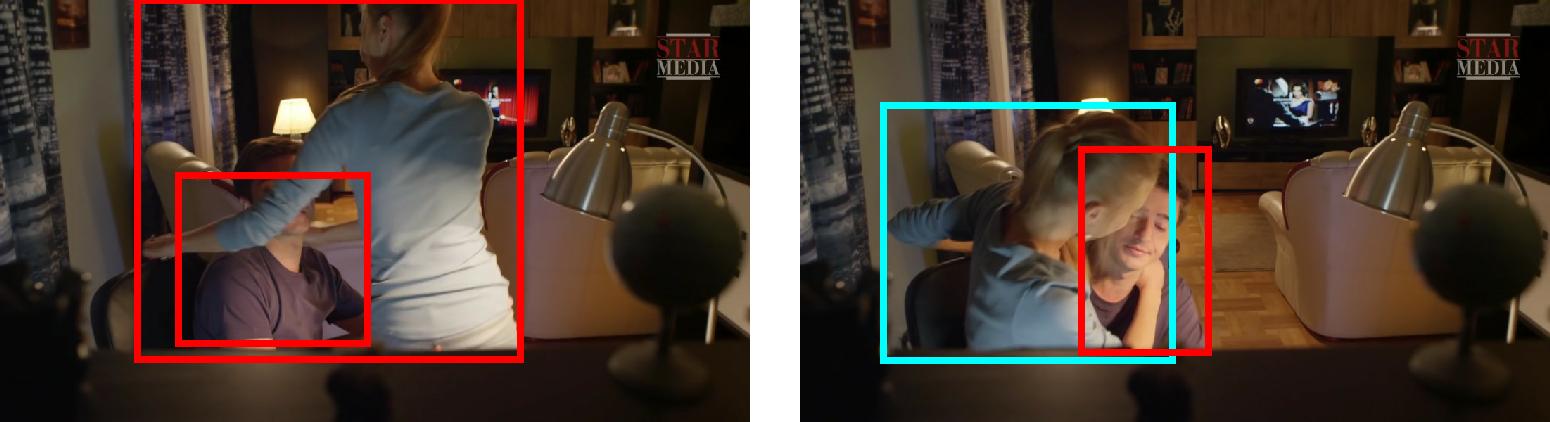} \\[-0.12cm]
grab (a person) $\rightarrow$ hug \\[0.11cm]\end{minipage}\hfill\begin{minipage}{0.32\linewidth}\centering
\includegraphics[width=\columnwidth]{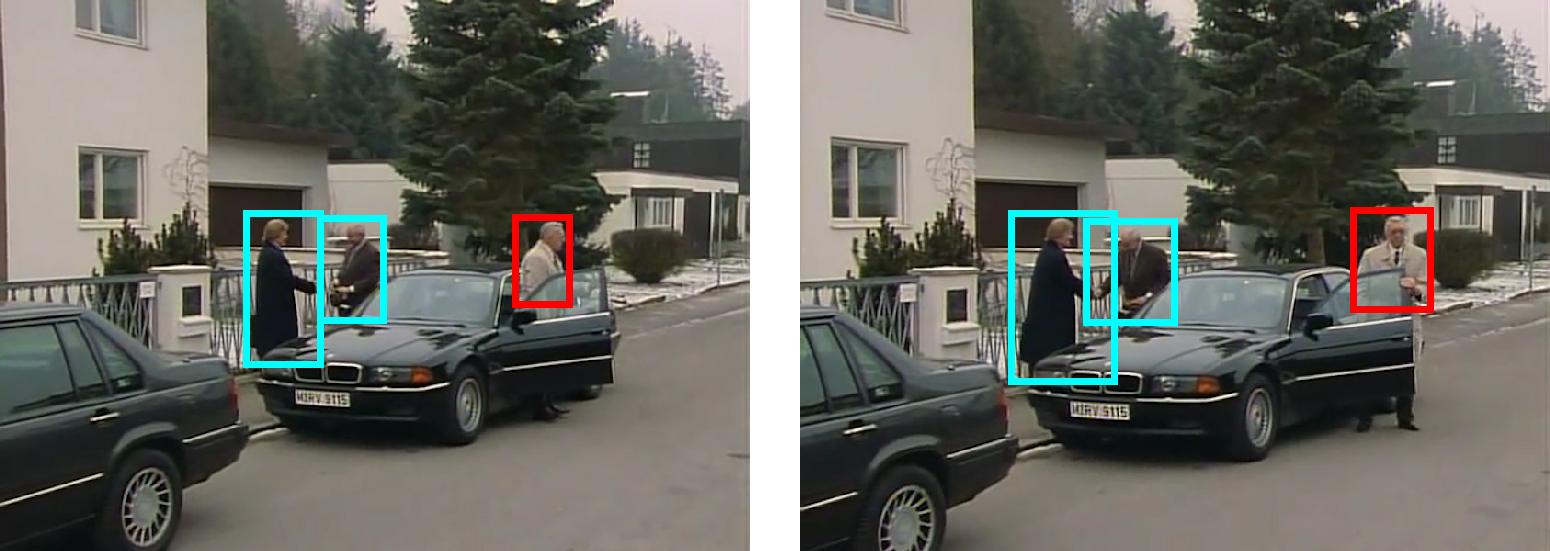} \\[-0.12cm]
open $\rightarrow$ close \\[0.11cm]
\includegraphics[width=\columnwidth]{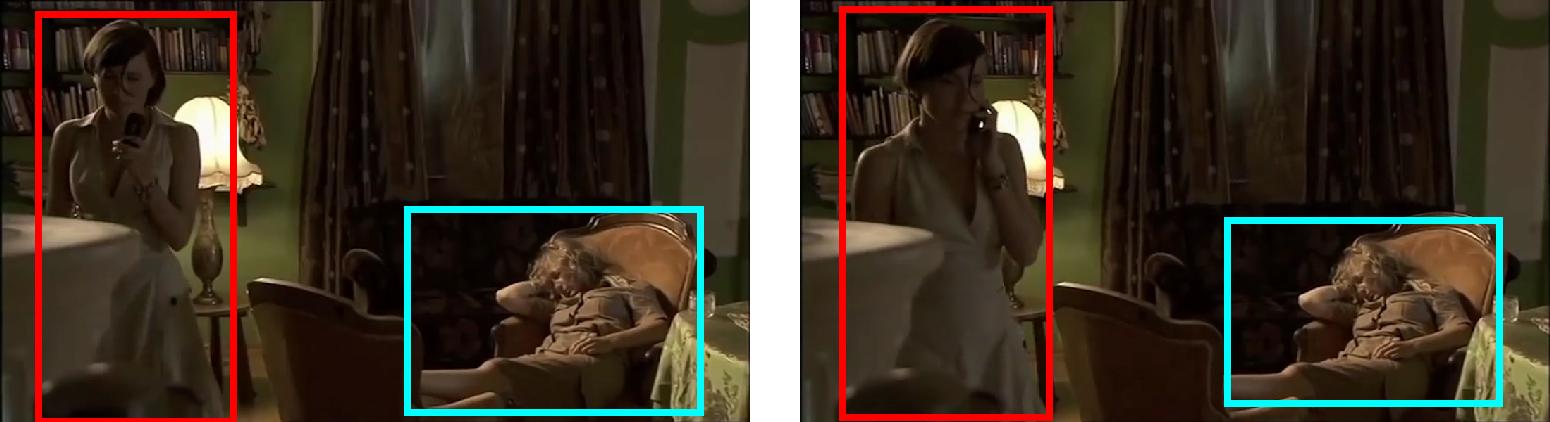} \\[-0.12cm]
look at phone $\rightarrow$ answer phone \\[0.11cm]\end{minipage}\hfill\begin{minipage}{0.32\linewidth}\centering
\includegraphics[width=\columnwidth]{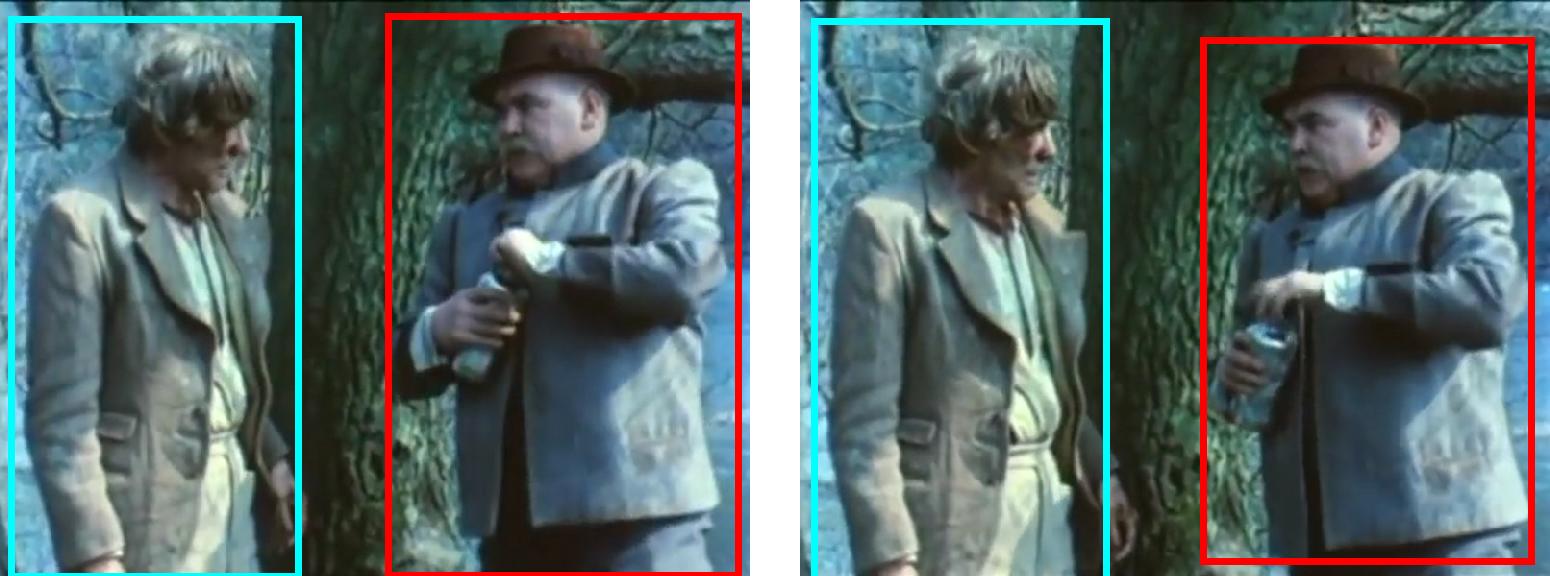} \\[-0.12cm]
turn $\rightarrow$ open \\[0.11cm]
\includegraphics[width=\columnwidth]{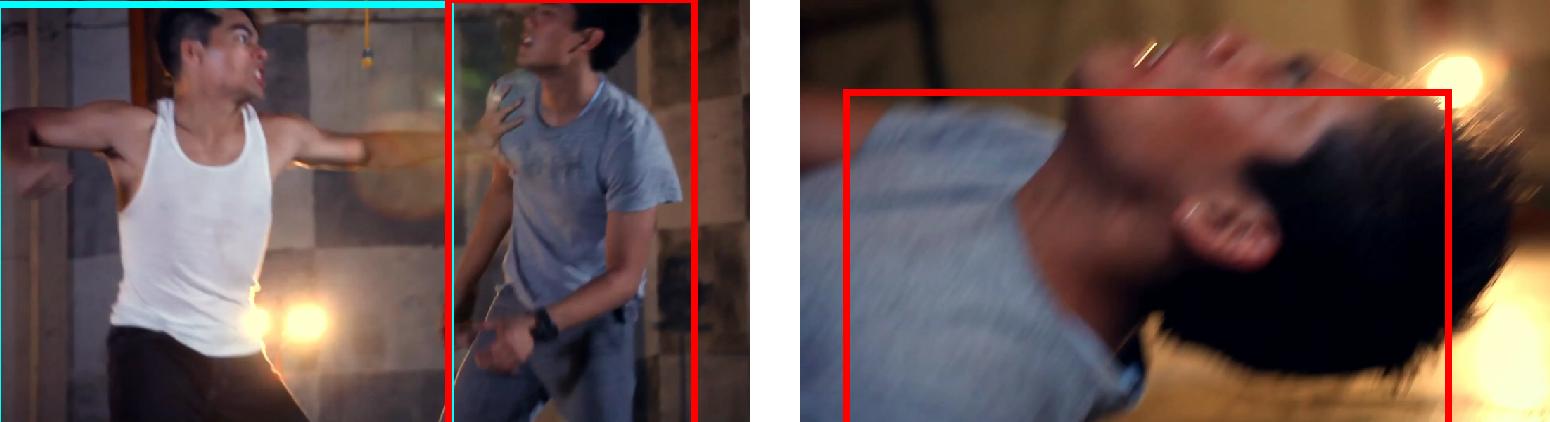} \\[-0.12cm]
fall down $\rightarrow$ lie/sleep \\[0.11cm]
\end{minipage}
\caption{We show examples of how atomic actions change over time in AVA. The text shows pairs of atomic actions for the people in \textcolor{red}{red} bounding boxes. Temporal information is key for recognizing many of the actions and appearance can substantially vary within an action category, such as opening a door or bottle.}\label{fig:example_annotations}
\vspace{-1em}
\end{figure*}

\subsection{Movie and segment selection}

The raw video content of the AVA dataset comes from YouTube. We begin by assembling a list of top actors of many different nationalities. For each name we issue a YouTube search query, retrieving up to 2000 results. We only include videos with the ``film'' or ``television'' topic annotation, a duration of over 30 minutes, at least 1 year since upload, and at least 1000 views. We further exclude black \& white, low resolution, animated, cartoon, and gaming videos, as well as those containing mature content.

To create a representative dataset within constraints, our selection criteria avoids filtering by action keywords, using automated action classifiers, or forcing a uniform label distribution. We aim to create an international collection of films by sampling from large film industries. However, the depiction of action in film is biased, e.g. by gender~\cite{geena_davis_2016}, and does not reflect the ``true'' distribution of human activity. 

Each movie contributes equally to the dataset, as we only label a sub-part ranging from the 15th to the 30th minute. We skip the beginning of the movie to avoid annotating titles or trailers. We choose a duration of 15 minutes so we are able to include more movies under a fixed annotation budget, and thus increase the diversity of our dataset. Each 15-min clip is then partitioned into 897 overlapping 3s movie segments with a stride of 1 second.

\subsection{Person bounding box annotation}

We localize a person and his or her actions with a bounding box. When multiple subjects are present in a keyframe, each subject is shown to the annotator separately for action annotation, and thus their action labels can be different.

Since bounding box annotation is manually intensive, we choose a hybrid approach. First, we generate an initial set of bounding boxes using the Faster-RCNN person detector~\cite{ren2015faster}. We set the operating point to ensure high-precision. Annotators then  annotate the remaining bounding boxes missed by our detector. This hybrid approach ensures full bounding box recall which is essential for benchmarking, while minimizing the cost of manual annotation. This manual annotation retrieves only 5\% more bounding boxes missed by our person detector, validating our design choice. Any incorrect bounding boxes are marked and removed by annotators in the next stage of action annotation.

\subsection{Person link annotation}

We link the bounding boxes over short periods of time to obtain ground-truth person tracklets. We calculate the pairwise similarity between bounding boxes in adjacent key frames using a person embedding~\cite{wu2016personnet} and solve for the optimal matching with the Hungarian algorithm~\cite{kuhn1955hungarian}. While automatic matching is generally strong, we further remove false positives with human annotators who verify each match. This procedure results in 81,000 tracklets ranging from a few seconds to a few minutes. 

\subsection{Action annotation}
\label{sec:action_label}

\begin{figure*}[tb!]
    \centering
    \includegraphics[width=1.9\columnwidth]{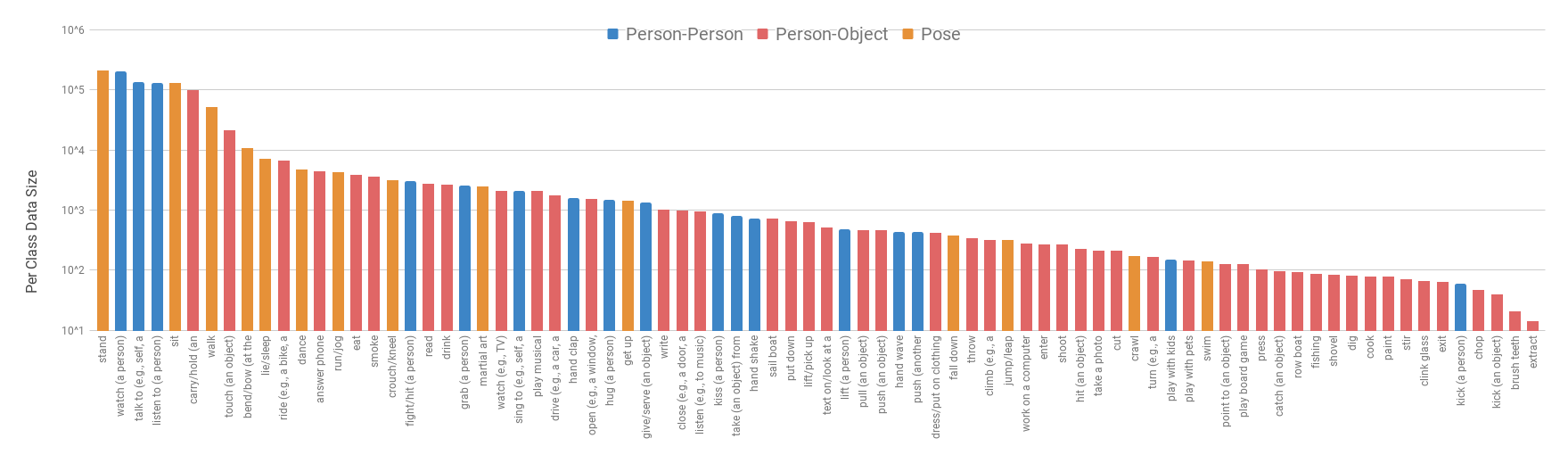}
    \vspace{-1em}
    \caption{Sizes of each action class in the AVA train/val dataset sorted by descending order, with colors indicating action types.}
    \label{fig:label_counts_top}
    \vspace{-1em}
\end{figure*}

The action labels are generated by crowd-sourced annotators using the interface shown in {Figure~\ref{fig:annotation}}. The left panel shows both the middle frame of the target segment (top) and the segment as a looping embedded video (bottom). The bounding box overlaid on the middle frame specifies the person whose action needs to be labeled. On the right are text boxes for entering up to 7 action labels, including 1 pose action (required), 3 person-object interactions (optional), and 3 person-person interactions (optional). If none of the listed actions is descriptive, annotators can flag a check box called ``other action''. In addition, they could flag segments containing blocked or inappropriate content, or incorrect bounding boxes. 

In practice, we observe that it is inevitable for annotators to miss correct actions when they are instructed to find all correct ones from a large vocabulary of 80 classes. Inspired by \cite{Sigurdsson_HCOMP2016}, we split the action annotation pipeline into two stages: action proposal and verification. We first ask multiple annotators to propose action candidates for each question, so the joint set possesses a higher recall than individual proposals. Annotators then verify these proposed candidates in the second stage. Results show significant recall improvement using this two-stage approach, especially on actions with fewer examples. See detailed analysis in the supplemental material. On average, annotators take 22 seconds to annotate a given video segment at the propose stage, and 19.7 seconds at the verify stage.  

Each video clip is annotated by three independent annotators and we only regard an action label as ground truth if it is verified by at least two annotators. Annotators are shown segments in randomized order.

\subsection{Training, validation and test sets}

Our training/validation/test sets are split at the video level, so that all segments of one video appear only in one split. The \numMovies{} videos are split into 235 training, 64 validation and 131 test videos, roughly a 55:15:30 split, resulting in 211k training, 57k validation and 118k test segments.

%% file: dataset.tex
\section{Characteristics of the AVA dataset}
\label{sec:stats}

We first build intuition on the diversity and difficulty of our AVA dataset through visual examples. Then, we characterize the annotations of our dataset quantitatively. Finally, we explore action and temporal structure.

\subsection{Diversity and difficulty}

Figure~\ref{fig:example_annotations} shows examples of atomic actions as they change over consecutive segments. Besides variations in bounding box size and cinematography, many of the categories will require discriminating fine-grained differences, such as ``clinking glass'' versus ``drinking'' or leveraging temporal context, such as ``opening'' versus ``closing''. 

Figure~\ref{fig:example_annotations} also shows two examples for the action ``open''. Even within an action class the appearance varies with vastly different contexts: the object being opened may even change.  The wide intra-class variety will allow us to learn features that identify the critical spatio-temporal parts of an action --- such as the breaking of a seal for ``opening''.

\subsection{Annotation Statistics}

\begin{figure}[tbp]
\centering
\includegraphics[width=\columnwidth]{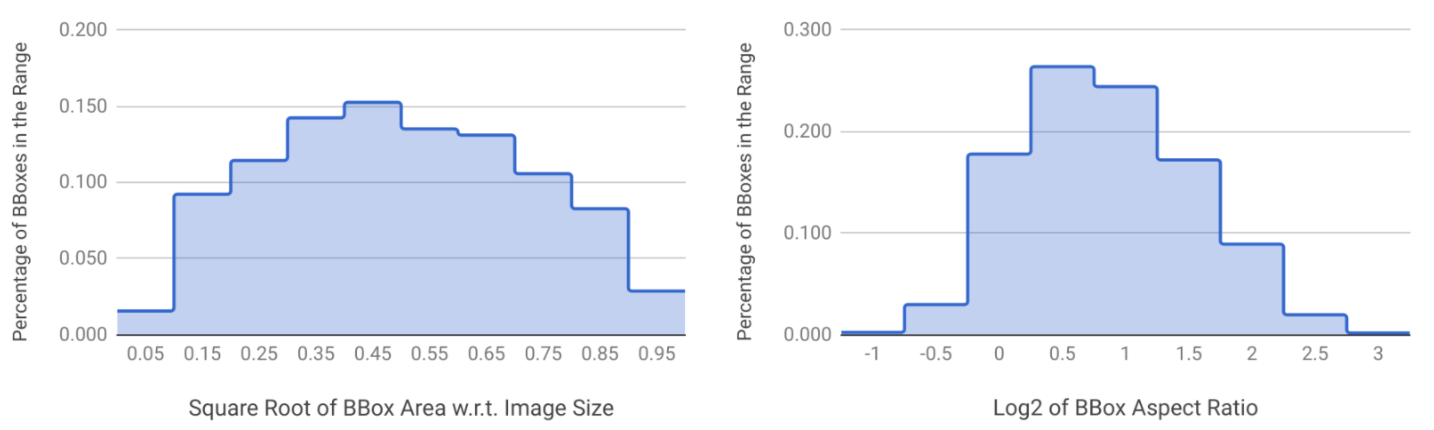}
\caption{Size and aspect ratio variations of annotated bounding boxes in the AVA dataset. Note that our bounding boxes consist of a large variation of sizes, many of which are small and hard to detect. Large variation also applies to the aspect ratios of bounding boxes, with mode at 2:1 ratio (e.g., sitting pose).}
\label{fig:bbox_stats}
\vspace{-1em}
\end{figure}

Figure~\ref{fig:label_counts_top} shows the distribution of action annotations in AVA. The distribution roughly follows Zipf's law. Figure~\ref{fig:bbox_stats} illustrates bounding box size distribution. A large portion of people take up the full height of the frame. However, there are still many boxes with smaller sizes. The variability can be explained by both zoom level as well as pose. For example, boxes with the label ``enter'' show the typical pedestrian aspect ratio of 1:2 with average widths of 30\% of the image width, and an average heights of 72\%. On the other hand, boxes labeled ``lie/sleep'' are close to square, with average widths of 58\% and heights of 67\%. The box widths are widely distributed, showing the variety of poses people undertake to execute the labeled actions. 

There are multiple labels for the majority of person bounding boxes. All bounding boxes have one pose label, 28\% of bounding boxes have at least 1 person-object interaction label, and 67\% of them have at least 1 person-person interaction label.

\subsection{Temporal Structure}
\label{sec:temp_structure}

A key characteristic of AVA is the rich temporal structure that evolves from segment to segment. Since we have linked people between segments, we can discover common consecutive actions by looking at pairs of actions performed by the same person. We sort pairs by Normalized Pointwise Mutual Information (NPMI)~\cite{Church1990}, which is commonly used in linguistics to represent the co-occurrence between two words: $\mathrm{NPMI}(x,y)=\left(\ln{\frac{p(x,y)}{p(x)p(y)}}\right) / \left(-\ln{p(x,y)}\right)$. Values intuitively fall in the range $(-1, 1]$, with $-1$ for pairs of words that never co-occur, $0$ for independent pairs, and $1$ for pairs that always co-occur.

\begin{table*}[tbp]
\begin{minipage}{0.47\linewidth}
    \centering
    \scalebox{0.9}{
    \begin{tabular}{l|l|c}
    First Action & Second Action & NPMI  \\
    \hline
ride (eg bike/car/horse) & drive (eg car/truck) & 0.68 \\
watch (eg TV) & work on a computer & 0.64 \\
drive (eg car/truck) & ride (eg car bike/car/horse) & 0.63\\
open (eg window/door) & close (eg door/box) & 0.59\\
text on/look at a cellphone & answer phone & 0.53\\
listen to (person) & talk to (person) & 0.47\\
fall down & lie/sleep & 0.46\\
talk to (person) & listen to (person) & 0.43\\
stand & sit & 0.40\\
walk & stand & 0.40\\
    \end{tabular}}
    \caption{We show top pairs of consecutive actions that are likely to happen before/after for the same person. We sort by NPMI.}
    \label{tab:temporal_npmi}
\end{minipage}\hfill\begin{minipage}{0.47\linewidth}
    \centering
    \scalebox{0.9}{
    \begin{tabular}{l|l|c}
    Person 1 Action & Person 2 Action & NPMI  \\
    \hline
ride (eg bike/car/horse) & drive (eg car/truck) & 0.60 \\
play musical instrument & listen (eg music) & 0.57 \\
take (object) & give/serve (object) & 0.51 \\
talk to (person) & listen to (person) & 0.46 \\
stand & sit & 0.31 \\
play musical instrument & dance & 0.23 \\
walk & stand & 0.21 \\
watch (person) & write & 0.15 \\
walk & run/jog & 0.15 \\
fight/hit (a person) & stand & 0.14 \\
    \end{tabular}}
    \caption{We show top pairs of simultaneous actions by different people. We sort by NPMI.}
    \label{tab:interperson_npmi}
\end{minipage}
\vspace{-1em}
\end{table*}

Table~\ref{tab:temporal_npmi} shows pairs of actions with top NPMI in consecutive one-second segments for the same person. After removing identity transitions, some interesting common sense temporal patterns arise. Frequently, there are transitions from ``look at phone''  $\rightarrow$ ``answer phone'', ``fall down''  $\rightarrow$ ``lie'', or ``listen to''  $\rightarrow$ ``talk to''.  We also analyze inter-person action pairs. Table \ref{tab:interperson_npmi} shows top pairs of actions performed at the same time, but by different people. Several meaningful pairs emerge, such as ``ride'' $\leftrightarrow$ ``drive'', ``play music'' $\leftrightarrow$ ``listen'', or ``take'' $\leftrightarrow$ ``give/serve''. The transitions between atomic actions, despite the relatively coarse temporal sampling, provide excellent data for building more complex models of actions and activities with longer temporal structure.
\vspace{-1em}

%% file: experiments.tex
\section{Action Localization Model} 
\label{sec:I3D}

Performance numbers on popular action recognition datasets such as UCF101 or JHMDB have gone up considerably in recent years, but we believe that this may present an artificially rosy picture of the state of the art. When the video clip involves only a single person performing something visually characteristic like swimming in an equally characteristic background scene, it is easy to classify accurately. Difficulties come in when actors are multiple, or small in image size, or performing actions which are only subtly different, and when the background scenes are not enough to tell us what is going on. AVA has these aspects galore, and we will find that performance at AVA is much poorer as a result. Indeed this finding was foreshadowed by the poor performance at the Charades dataset~\cite{charades2016}.

\begin{figure}[b]
\vspace{-2em}
\centering
\includegraphics[width=0.48\textwidth]{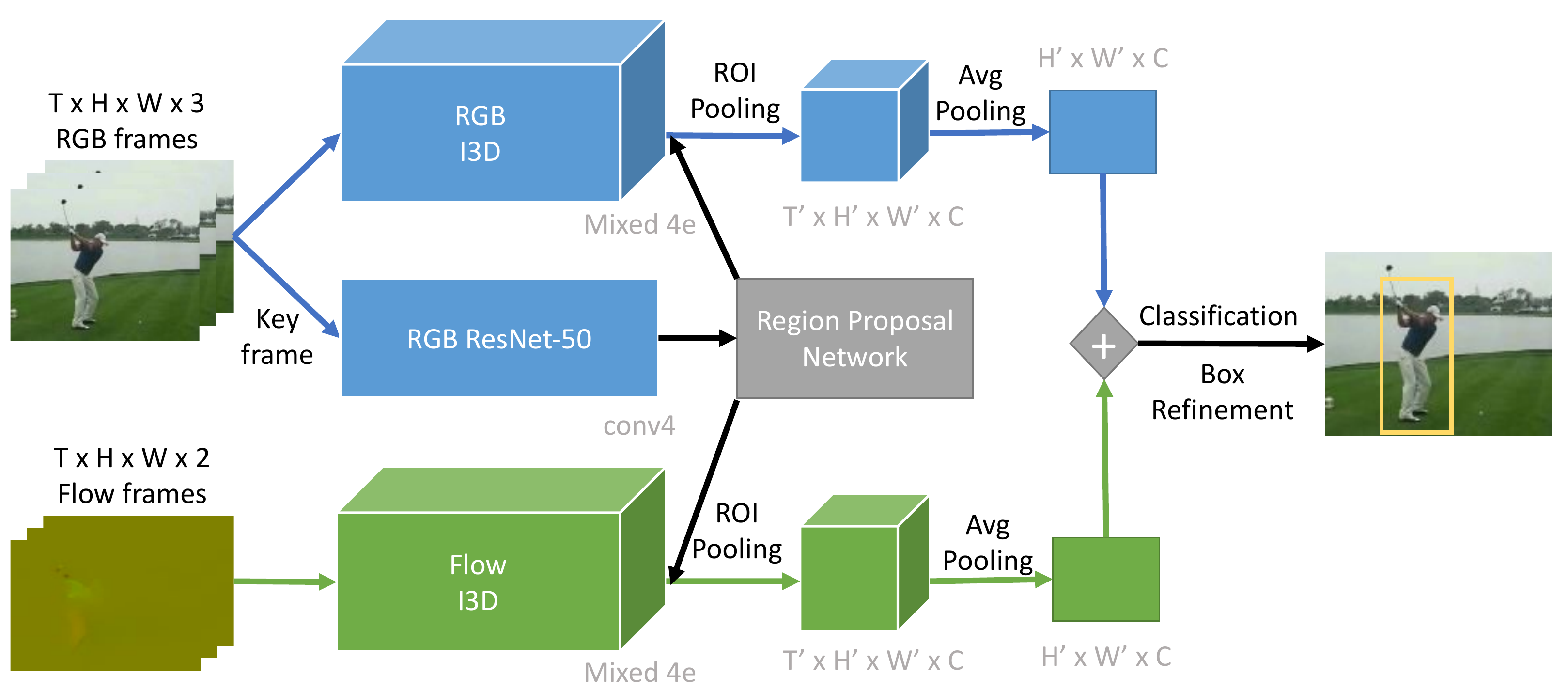}
\caption{Illustration of our approach for spatio-temporal action localization. Region proposals are detected and regressed with Faster-RCNN on RGB keyframes. Spatio-temporal tubes are classified with two-stream I3D convolutions. 
}
\label{fig:model_figure}
\end{figure}

To prove our point, we develop a state of the art action localization approach inspired by recent approaches for spatio-temporal action localization that operate on multi-frame temporal information~\cite{T_CNN_iccv17,tubelets_iccv17}. Here, we rely on the impact of larger temporal context based on I3D~\cite{i3d_cvpr17} for action detection. See Fig.~\ref{fig:model_figure} for an overview of our approach. 

Following Peng and Schmid~\cite{peng2016multi}, we apply the Faster RCNN algorithm~\cite{ren2015faster} for end-to-end localization and classification of actions. However, in their approach, the temporal information is lost at the first layer where input channels from multiple frames are concatenated over time. We propose to use the Inception 3D (I3D) architecture by Carreira and Zisserman~\cite{i3d_cvpr17} to model temporal context. The I3D architecture is designed based on the Inception architecture~\cite{szegedy2015rethinking}, but replaces 2D convolutions with 3D convolutions. Temporal information is kept throughout the network. I3D achieves state-of-the-art performance on a wide range of video classification benchmarks.

To use I3D with Faster RCNN, we make the following changes to the model: first, we feed input frames of length $T$ to the I3D model, and extract 3D feature maps of size $T' \times W' \times H' \times C$ at the \textit{Mixed 4e} layer of the network. The output feature map at \textit{Mixed 4e} has a stride of 16, which is equivalent to the conv4 block of ResNet~\cite{he2016resnet}. Second, for action proposal generation, we use a 2D ResNet-50 model on the keyframe as the input for the region proposal network, avoiding the impact of I3D with different input lengths on the quality of generated action proposals. Finally, we extend ROI Pooling to 3D by applying the 2D ROI Pooling at the same spatial location over all time steps. To understand the impact of optical flow for action detection, we fuse the RGB stream and the optical flow stream at the feature map level using average pooling.

\noindent {\bf Baseline.}
To compare to a frame-based two-stream approach on AVA, we implement a variant of~\cite{peng2016multi}. We  use Faster RCNN~\cite{ren2015faster} with ResNet-50~\cite{he2016resnet} to jointly learn action proposals and action labels. Region proposals are obtained with the RGB stream only. 
The region classifier takes as input RGB along with optical flow features stacked over 5 consecutive frames. As for our I3D approach, we jointly train the RGB and the optical flow streams by fusing the conv4 feature maps with average pooling.  

\noindent {\bf Implementation details.} We implement FlowNet v2~\cite{flownet2} to extract optical flow features.  We train Faster-RCNN with asynchronous SGD. For all training tasks, we use a validation set to determine the number of training steps, which ranges from 600K to 1M iterations. We fix the input resolution to be 320 by 400 pixels. All the other model parameters are set based on the recommended values from~\cite{huang2016coco}, which were tuned for object detection. The ResNet-50 networks are initialized with ImageNet pre-trained models. For the optical flow stream, we duplicate the conv1 filters to input 5 frames. The I3D networks are initialized with Kinetics~\cite{kinetics17} pre-trained models, for both the RGB and optical flow streams. Note that although I3D were pre-trained on 64-frame inputs, the network is fully convolutional over time and can take any number of frames as input. All feature layers are jointly updated during training. The output frame-level detections are post-processed with non-maximum suppression with threshold 0.6.

One key difference between AVA and existing action detection datasets is that the action labels of AVA are not mutually exclusive. To address this, we replace the standard softmax loss function by a sum of binary Sigmoid losses, one for each class. We use Sigmoid loss for AVA and softmax loss for all other datasets.

\noindent {\bf Linking.}
Once we have per frame-level detections,
we link them to construct action tubes. We report video-level performance based on average scores over the obtained tubes. We use the same linking algorithm as described in \cite{Singh_ICCV2017}, except that we do not apply temporal labeling. Since AVA is annotated at 1~Hz and each tube may have multiple labels, we modify the video-level evaluation protocol to estimate an upper bound. We use ground truth links to infer detection links, and when computing IoU score of a class between a ground truth tube and a detection tube, we only take tube segments that are labeled by that class into account.
\vspace{-0.5em}

\section{Experiments and Analysis}
\label{sec:experiments}

We now experimentally analyze key characteristics of AVA and motivate challenges for action understanding.
\vspace{-0.5em}

\subsection{Datasets and Metrics} 

\noindent \textbf{AVA benchmark.}
Since the label distribution in AVA roughly follows Zipf's law (Figure~\ref{fig:label_counts_top}) and evaluation on a very small number of examples could be unreliable, we use classes that have at least 25 instances in validation and test splits to benchmark performance. Our resulting benchmark consists of a total of 210,634 training, 57,371 validation and 117,441 test examples on 60 classes. Unless otherwise mentioned, we report results trained on the training set and evaluated on the validation set. We randomly select 10\% of the training data for model parameter tuning.

\noindent {\bf Datasets.} Besides AVA, we also analyze standard video datasets in order to compare difficulty. JHMDB~\cite{jhmdb} consists of 928 trimmed clips over 21 classes. We report results for split one in our ablation study, but results are averaged over three splits for comparison to the state of the art. For UCF101, we use spatio-temporal annotations for a 24-class subset with 3207 videos, provided by Singh \etal~\cite{Singh_ICCV2017}. We conduct experiments on the official split1 as is standard.

\begin{figure*}[tb]
\centering
\includegraphics[width=\textwidth]{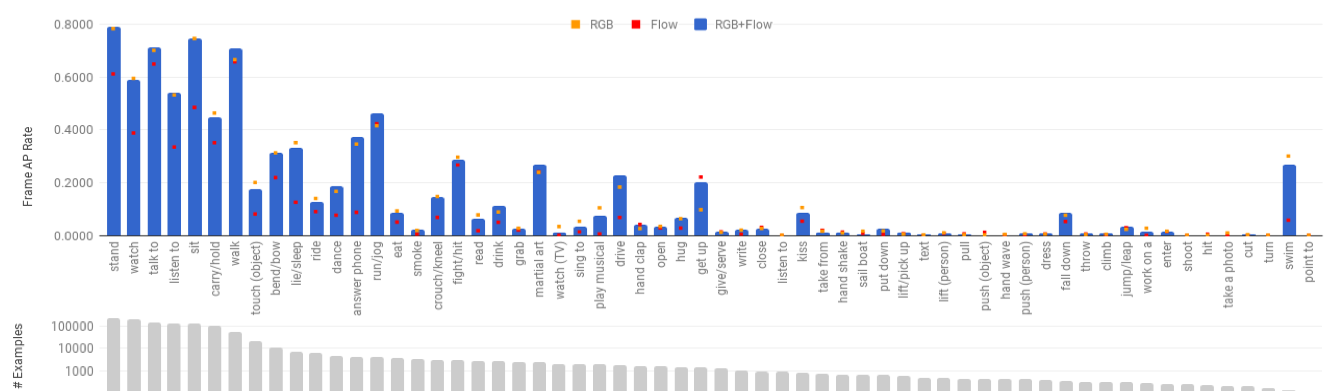}
\caption{Top: We plot the performance of models for each action class, sorting by the number of training examples. Bottom: We plot the number of training examples per class. While more data is better, the outliers suggest that not all classes are of equal complexity.  For example, one of the smallest classes ``swim'' has one of the highest performances because the associated scenes make it relatively easy.}
%Instead, the challenging categories are ones with large diversity, such as ``touch,'' where context is not as discriminative.}
\label{fig:data_vs_ap}
\vspace{-1em}
\end{figure*}

\noindent {\bf Metrics.} For evaluation, we follow standard practice when possible. We report intersection-over-union (IoU) performance on frame level and video level. For frame-level IoU, we follow the standard protocol used by the PASCAL VOC challenge~\cite{PASCAL} and report the average precision (AP) using an IoU threshold of 0.5. For each class, we compute the average precision and report the average over all classes. For video-level IoU, we compute 3D IoUs between ground truth tubes and linked detection tubes at the threshold of 0.5. The mean AP is computed by averaging over all classes.

\begin{table}[tb]
\centering
\begin{tabular}{p{3.0cm}|c|c}
\parbox[l]{2.9cm}{Frame-mAP} & \parbox[c]{1.2cm}{JHMDB} & \parbox[c]{1.85cm}{UCF101-24} \\
\hline
Actionness~\cite{actionness_cvpr16} & 39.9\% & - \\
Peng w/o MR~\cite{peng2016multi} & 56.9\% & 64.8\% \\
Peng w/ MR~\cite{peng2016multi} & 58.5\% & 65.7\% \\
ACT~\cite{tubelets_iccv17} & 65.7\% & 69.5\% \\
\hline
Our approach & \textbf{73.3\%} & \textbf{76.3\%} \\
\end{tabular}
\\[0.5em]
\centering
\begin{tabular}{p{3.0cm}|c|c}
\parbox[l]{2.9cm}{Video-mAP} & \parbox[c]{1.2cm}{JHMDB} & \parbox[c]{1.85cm}{UCF101-24} \\\hline
Peng w/ MR~\cite{peng2016multi} & 73.1\% & 35.9\% \\
Singh~\etal~\cite{Singh_ICCV2017} & 72.0\% & 46.3\% \\
ACT~\cite{tubelets_iccv17} & 73.7\% & 51.4\%  \\
TCNN~\cite{T_CNN_iccv17} & 76.9\% & - \\
\hline
Our approach & \textbf{78.6\%} & \textbf{59.9\%} \\
\end{tabular}
\caption{Frame-mAP (top) and video-mAP (bottom) @ IoU 0.5 for JHMDB and UCF101-24. For JHMDB, we report averaged performance over three splits. Our approach outperforms previous state-of-the-art on both metrics by a considerable margin.}
\label{tab:overall_comparison}
\vspace{-1em}
\end{table}

\subsection{Comparison to the state-of-the-art}

Table~\ref{tab:overall_comparison} shows our model performance on two standard video datasets. Our 3D two-stream model obtains state-of-the-art performance on UCF101 and JHMDB, outperforming well-established baselines for both frame-mAP and video-mAP metrics.

However, the picture is less auspicious when recognizing atomic actions. Table~\ref{tab:exp_ablations} shows that the same model obtains relatively low performance on AVA validation set (frame-mAP of \textbf{15.6\%}, video-mAP of \textbf{12.3\%} at 0.5 IoU and \textbf{17.9\%} at 0.2 IoU), as well as test set (frame-mAP of \textbf{14.7\%}). We attribute this to the design principles behind AVA: we collected a vocabulary where context and object cues are not as discriminative for action recognition. Instead, recognizing fine-grained details and rich temporal models may be needed to succeed at AVA, posing a new challenge for visual action recognition. In the remainder of this paper, we analyze what makes AVA challenging and discuss how to move forward. 

\subsection{Ablation study}

\noindent \textbf{How important is temporal information for recognizing AVA categories?} Table~\ref{tab:exp_ablations} shows the impact of the temporal length and the type of model. All 3D models outperform the 2D baseline on JHMDB and UCF101-24. For AVA, 3D models perform better after using more than 10 frames. We can also see that increasing the length of the temporal window helps for the 3D two-stream models across all datasets. As expected, combining RGB and optical flow features improves the performance over a single input modality.  Moreover, AVA benefits more from larger temporal context than JHMDB and UCF101, whose performances saturate at 20 frames. This gain and the consecutive actions in Table \ref{tab:temporal_npmi} suggests that one may obtain further gains by leveraging the rich temporal context in AVA. 

\begin{table}
\footnotesize
\centering
\begin{tabular}{c|c|c|c|c}
Model & Temp.+ Mode & JHMDB & UCF101-24 & AVA \\
\hline
2D & 1 RGB + 5 Flow & 52.1\% & 60.1\% & 13.7\% \\
3D & 5 RGB + 5 Flow & 67.9\% & 76.1\% & 13.6\% \\
3D & 10 RGB + 10 Flow & 73.4\% & 78.0\% & 14.6\% \\
3D & 20 RGB + 20 Flow & 76.4\% & 78.3\% & 15.2\% \\
3D & 40 RGB + 40 Flow & 76.7\% & 76.0\% & \textbf{15.6}\% \\
3D & 50 RGB + 50 Flow & - & 73.2\% & 15.5\% \\
\hline
3D & 20 RGB & 73.2\% & 77.0\% & 14.5\% \\
3D & 20 Flow & 67.0\% & 71.3\% & 9.9\% \\
\end{tabular}
\caption{Frame-mAP @ IoU 0.5 for action detection on JHMDB (split1), UCF101 (split1) and AVA. Note that JHMDB has up to 40 frames per clip. For UCF101-24, we randomly sample 20,000 frame subset for evaluation. Although our model obtains state-of-the-art performance on JHMDB and UCF101-24, the fine-grained nature of AVA makes it a challenge.}
\label{tab:exp_ablations}
\vspace{-1em}
\end{table}

\begin{figure}
    \centering
    \includegraphics[width=\linewidth]{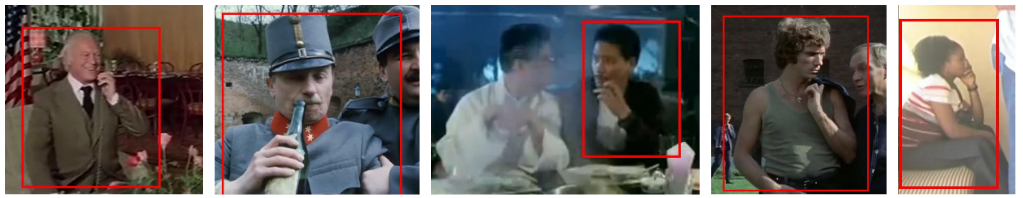}
    \caption{Red boxes show high-scoring false alarms for smoking. The model often struggles to discriminate fine-grained details.}
    \label{fig:falsealarms}
    \vspace{-1em}
\end{figure}

\noindent \textbf{How challenging is localization versus recognition?} Table~\ref{tab:localization_comparison} compares the performance of end-to-end action localization and recognition versus class agnostic action localization. We can see that although action localization is more challenging on AVA than on JHMDB, the gap between localization and end-to-end detection performance is nearly 60\% on AVA, while less than 15\% on JHMDB and UCF101. This suggests that the main difficulty of AVA lies in action classification rather than localization.  Figure~\ref{fig:falsealarms} shows examples of high-scoring false alarms, suggesting that the difficulty in recognition lies in the fine-grained details. 
 
\begin{table}[tbp]
\centering
\begin{tabular}{l|c|c|c}
& JHMDB & UCF101-24 & AVA \\
\hline
Action detection & 76.7\% & 76.3\% & 15.6\% \\
Actor detection & 92.8\% & 84.8\% & 75.3\% \\
\end{tabular}
\caption{Frame-mAP @ IoU 0.5  for action detection and actor detection performance on JHMDB (split1), UCF101-24 (split1) and AVA benchmarks. Since human annotators are consistent, our results suggest there is significant headroom to improve on recongizing atomic visual actions.}
\label{tab:localization_comparison}
\vspace{-1em}
\end{table}

\noindent \textbf{Which categories are challenging? How important is number of training examples?}  Figure~\ref{fig:data_vs_ap} breaks down performance by categories and the number of training examples. While more data generally yields better performance, the outliers reveals that not all categories are of equal complexity. Categories correlated with scenes and objects (such as swimming) or categories with low diversity (such as fall down) obtain high performance despite having fewer training examples. In contrast, categories with lots of data, such as touching and smoking, obtain relatively low performance possibly because they have large visual variations or require fine grained discrimination, motivating work on person-object interaction \cite{hico2015,something_iccv17}. We hypothesize that the gains on recognizing atomic actions will need not only large datasets, such as AVA, but also rich models of motion and interactions.

%% file: conclusion.tex
\vspace{-1mm}
\section{Conclusion}
\label{sec:conclusion}

This paper introduces the AVA dataset with spatio-temporal annotations of atomic actions at 1~Hz over diverse 15-min.\ movie segments. In addition we propose a method that outperforms the current state of the art on standard benchmarks to serve as a baseline. This method highlights the difficulty of the AVA dataset as its performance is significantly lower than on UCF101 or JHMDB, underscoring the need for developing new action recognition approaches.

Future work includes modeling more complex activities based on our atomic actions. Our present day visual classification technology may enable us to classify events such as ``eating in a restaurant'' at the coarse scene/video level, but models based on AVA's fine spatio-temporal granularity facilitate understanding at the level of an individual agent’s actions. These are essential steps towards imbuing computers with ``social visual intelligence'' -- understanding what humans are doing, what they might do next, and what they are trying to achieve.

%% file: supplemental.tex
\begin{center}
    %\large\bf Supplementary Material
    \large\bf Appendix
\end{center}
%\section*{Appendix}
%\label{sec:supplemental}

\noindent
In the following, we present additional quantitative information and examples for our AVA dataset as well as for our action detection approach on AVA. 

\section{Additional details on the annotation}

\begin{figure}[h]
    \centering
    \includegraphics[width=0.95\columnwidth]{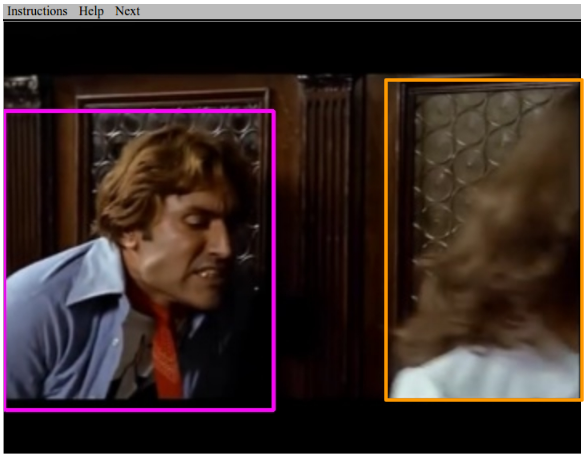}
    \caption{User interface for bounding box annotation. The purple box was generated by the person detector. The orange box (missed by the detector) was manually added by an annotator. }
    \label{fig:ui_bbox}
\end{figure}

Figure~\ref{fig:ui_bbox} shows the user interface for bounding box annotation. As described in Section 3.3, we employ a hybrid approach to tradeoff accuracy with annotation cost. We show annotators frames overlaid by detected person bounding boxes, so they can add boxes to include more persons missed by the detector.

\begin{figure*}
\centering
\includegraphics[width=\linewidth]{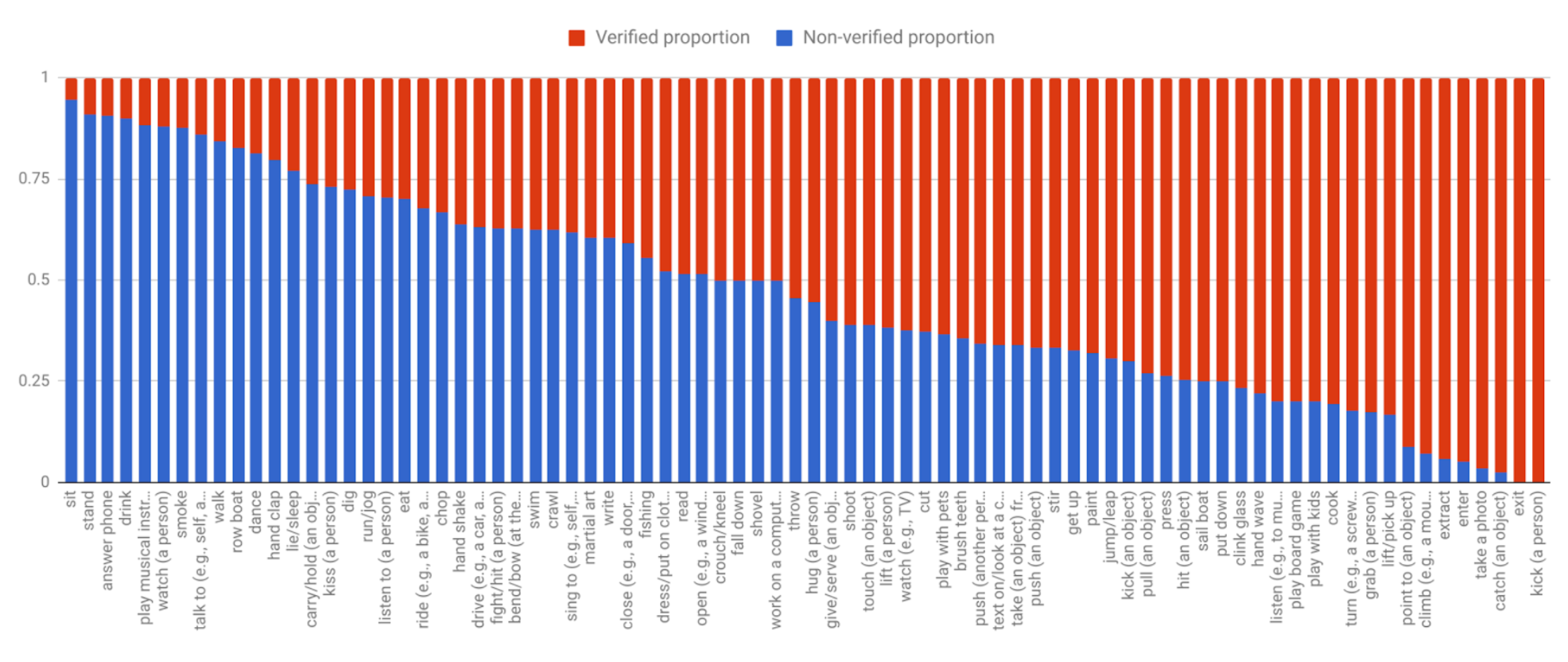}
\caption{Action class recall improvement due to the two-stage process. For each class, the blue bar shows the proportion of labels annotated without verification (majority voted results over raters' selections from 80 classes.), and the red bar shows the proportion of labels revived from the verification stage. More than half of the action classes doubles their recalls thanks to the additional verification.}
\label{fig:verify_recall}
\end{figure*}

In Section 3.5 of our paper submission, we explain why our two-stage action annotation design is crucial for preserving high recall of action classes. Here we show quantitative analysis. Figure~\ref{fig:verify_recall} shows the proportion of labels per action class generated from each stage. (Blue ones are generated from the first (propose) stage and red ones from the second (verify) stage). As we can see, for more than half of our action labels, the majority labels are derived from the verification stage. Furthermore, the smaller the action class size, the more likely that they are missed by the first stage (e.g., kick, exit, extract), and require the second stage to boost recall. The second stage helps us to build more robust models for long tail classes that are more sensitive to the sizes of the training data.

\section{Additional details on the dataset}

%The attached video includes a sequence of 3-second video segments for a subset of AVA action classes. Each class is illustrated by two example segments with bounding boxes annotated in the middle frames.

Table~\ref{tab:label_counts_all1} and ~\ref{tab:label_counts_all2} present the number of instances for each class of the AVA trainval dataset. We observe a significant class imbalance to be expected in real-world data [c.f.\ Zipf's Law]. As stated in the paper, we select a subset of these classes (without asterisks) for our benchmarking experiment, in order to have a sufficient number of test examples. Note that we consider the presence of the ``rare'' classes as an opportunity for approaches to learn from a few training examples. 

\begin{table*}[h]
    \centering
    \small
    \begin{tabular}{cc}    
    \begin{tabular}{l|c}
    Pose & \# \\
    \hline
stand & 208332  \\ 
sit & 127917  \\ 
walk & 52644  \\ 
bend/bow (at the waist) & 10558  \\ 
lie/sleep & 7176  \\ 
dance & 4672  \\ 
run/jog & 4293  \\ 
crouch/kneel & 3125  \\ 
martial art & 2520  \\ 
get up & 1439  \\ 
fall down & 378  \\
jump/leap & 313  \\
crawl* & 169  \\
swim & 141  \\ 
&\\
&\\
&\\
\end{tabular} &
    \begin{tabular}{l|c}
    Person-Person Interaction & \# \\
    \hline
watch (a person) & 202369  \\
talk to (e.g. self/person) & 136780  \\ 
listen to (a person) & 132174  \\ 
fight/hit (a person) & 3069  \\
grab (a person)	& 2590  \\
sing to (e.g., self, a person, a group) & 2107  \\
hand clap & 1539  \\
hug (a person) & 1472  \\
give/serve (an object) to (a person) & 1337  \\
kiss (a person) & 898  \\
take (an object) from (a person) & 783  \\
hand shake & 719  \\
lift (a person) & 481  \\
push (another person) & 434  \\
hand wave & 430  \\
play with kids* & 151  \\
kick (a person)* & 60  \\
\end{tabular} 
\\
\end{tabular}
    \caption{Number of instances for pose (left) and person-person (right) interaction labels in the AVA trainval dataset, sorted in decreasing order. Labels marked by asterisks are not included in the benchmark dataset.}
    \label{tab:label_counts_all1}
\end{table*}

\begin{table*}[h]
    \centering
    \small
    \begin{tabular}{cc}   
    \begin{tabular}{l|c}
    Person-Object Interaction & \# \\
    \hline
carry/hold (an object) & 100598  \\
touch (an object) & 21099  \\
ride (e.g., a bike, a car, a horse) & 6594  \\
answer phone & 4351  \\
eat & 3889  \\
smoke & 3528  \\
read & 2730  \\
drink & 2681  \\
watch (e.g., TV) & 2105  \\
play musical instrument & 2063  \\
drive (e.g., a car, a truck) & 1728  \\
open (e.g., a window, a car door) & 1547  \\
write & 1014  \\
close (e.g., a door, a box) & 986  \\
listen (e.g., to music) & 950  \\
sail boat & 699  \\
put down & 653  \\
lift/pick up & 634  \\
text on/look at a cellphone & 517  \\
push (an object) & 465  \\
pull (an object) & 460  \\
dress/put on clothing & 420  \\
throw & 336  \\
climb (e.g., a mountain) & 315  \\
work on a computer & 278  \\
enter & 271  \\
\end{tabular} &
    \begin{tabular}{l|c}
    Person-Object Interaction & \# \\
    \hline
shoot & 267  \\
hit (an object) & 220  \\
take a photo & 213  \\
cut & 212  \\
turn (e.g., a screwdriver) & 167  \\
play with pets* & 146  \\
point to (an object) & 128  \\
play board game* & 127  \\
press* & 102  \\
catch (an object)* & 97  \\
fishing* & 88  \\
cook* & 79  \\
paint* & 79  \\
shovel* & 79  \\
row boat* & 77  \\
dig* & 72  \\
stir* & 71  \\
clink glass* & 67  \\
exit* & 65  \\
chop* & 47  \\
kick (an object)* & 40  \\
brush teeth* & 21  \\
extract* & 13  \\
&\\
\end{tabular}
\\
\end{tabular}
    \caption{Number of instances for person-object interactions in the AVA trainval dataset, sorted in decreasing order. Labels marked by asterisks are not included in the benchmark.}
    \label{tab:label_counts_all2}
\end{table*}

Figure \ref{fig:pairs} shows more examples of common consecutive atomic actions in AVA.

\begin{figure*}[t]
\centering
\includegraphics[width=0.3\linewidth]{pairs_sup/./answerphone_lookatphone/1f298a0b34096c1f_1360000000_1361000000.jpg}\hfill
\includegraphics[width=0.3\linewidth]{pairs_sup/./answerphone_lookatphone/5d5fc5177582ee40_1273000000_1274000000.jpg}\hfill
\includegraphics[width=0.3\linewidth]{pairs_sup/./answerphone_lookatphone/c499913595430c26_1165000000_1166000000.jpg}
\\
answer (eg phone) $\rightarrow$ look at (eg phone)
\noindent\rule{\textwidth}{1pt}\\[1em]
\includegraphics[width=0.3\linewidth]{pairs_sup/./answerphone_putdown/31ada18139af78a4_1708000000_1709000000.jpg}\hfill
\includegraphics[width=0.3\linewidth]{pairs_sup/./answerphone_putdown/8237608f4e05cd1d_1407000000_1408000000.jpg}\hfill
\includegraphics[width=0.3\linewidth]{pairs_sup/./answerphone_putdown/df61d1dcc9c367c8_1041000000_1042000000.jpg}
\\
answer (eg phone) $\rightarrow$ put down
\noindent\rule{\textwidth}{1pt}\\[1em]
\includegraphics[width=0.3\linewidth]{pairs_sup/./clinkglass_drink/01879b5d0f1e5243_1192000000_1193000000.jpg}\hfill
\includegraphics[width=0.3\linewidth]{pairs_sup/./clinkglass_drink/11c8afa7cb7731d6_1356000000_1357000000.jpg}\hfill
\includegraphics[width=0.3\linewidth]{pairs_sup/./clinkglass_drink/1ca8d1ef418244f1_1004000000_1005000000.jpg}
\\
clink glass $\rightarrow$ drink
\noindent\rule{\textwidth}{1pt}\\[1em]
\includegraphics[width=0.3\linewidth]{pairs_sup/./crouch_crawl/0fe0494d4e8dc59f_1757000000_1758000000.jpg}\hfill
\includegraphics[width=0.3\linewidth]{pairs_sup/./crouch_crawl/1afa7e723ddb9886_1316000000_1317000000.jpg}\hfill
\includegraphics[width=0.3\linewidth]{pairs_sup/./crouch_crawl/29bd5f76e8fe8dd6_1283000000_1284000000.jpg}
\\
crouch/kneel $\rightarrow$ crawl
\noindent\rule{\textwidth}{1pt}\\[1em]
\includegraphics[width=0.3\linewidth]{pairs_sup/./grab_handshake/1c9ce027d663bc99_1310000000_1311000000.jpg}\hfill
\includegraphics[width=0.3\linewidth]{pairs_sup/./grab_handshake/cfe7ec2e9187abaa_1436000000_1437000000.jpg}\hfill
\includegraphics[width=0.3\linewidth]{pairs_sup/./grab_handshake/d8fa71886d16535f_1136000000_1137000000.jpg}
\\
grab $\rightarrow$ handshake
\noindent\rule{\textwidth}{1pt}\\[1em]
\includegraphics[width=0.3\linewidth]{pairs_sup/./grab_hug/2786ede32f528530_1202000000_1203000000.jpg}\hfill
\includegraphics[width=0.3\linewidth]{pairs_sup/./grab_hug/7abede7a2241e9d2_1288000000_1289000000.jpg}\hfill
\includegraphics[width=0.3\linewidth]{pairs_sup/./grab_hug/e410e3d28c398670_1042000000_1043000000.jpg}
\\
grab $\rightarrow$ hug
\noindent\rule{\textwidth}{1pt}\\[1em]
\includegraphics[width=0.3\linewidth]{pairs_sup/./open_close/20a7412dc8aeffe0_1467000000_1468000000.jpg}\hfill
\includegraphics[width=0.3\linewidth]{pairs_sup/./open_close/94398b716581a751_1781000000_1782000000.jpg}\hfill
\includegraphics[width=0.3\linewidth]{pairs_sup/./open_close/ab1daf00ee687e6a_1092000000_1093000000.jpg}
\\
open $\rightarrow$ close\\[1em]
\caption{We show more examples of how atomic actions change over time in AVA. The text shows pairs of atomic actions for the people in \textcolor{red}{red} bounding boxes.}
\label{fig:pairs}
\end{figure*}

\section{Examples of our action detection}

\begin{figure*}[t]
\centering
\includegraphics[width=0.8\linewidth]{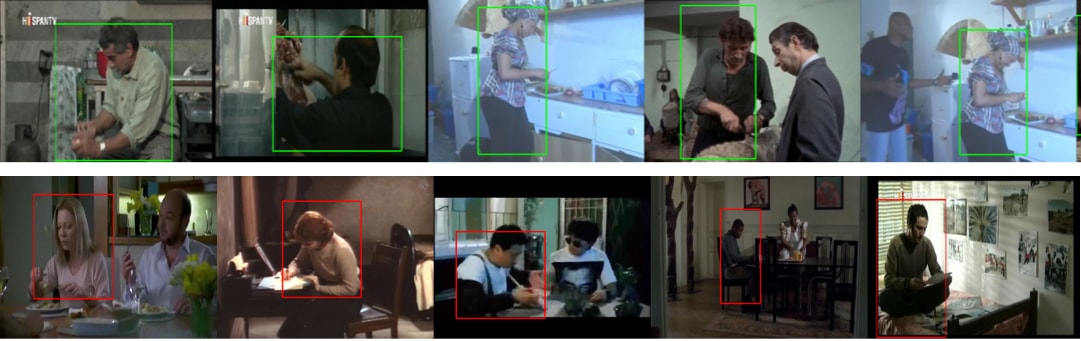}
\\
cut
\\
\noindent\rule{.85\textwidth}{1pt}\\[1em]
\includegraphics[width=0.8\linewidth]{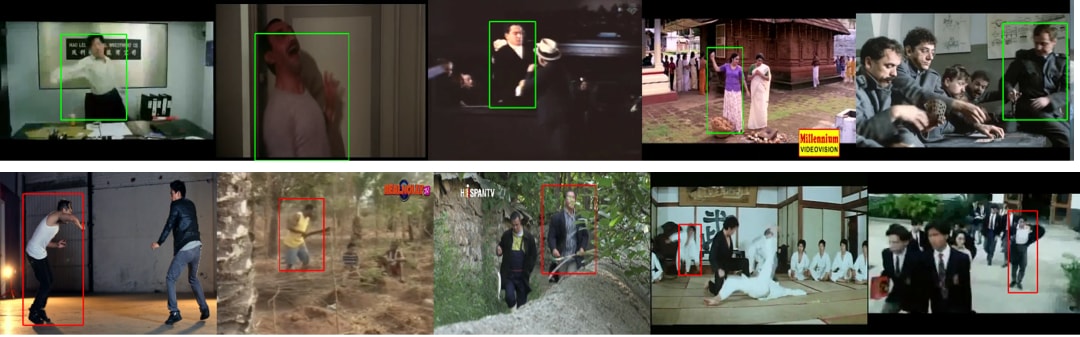}
\\
throw
\\
\noindent\rule{.85\textwidth}{1pt}\\[1em]
\includegraphics[width=0.8\linewidth]{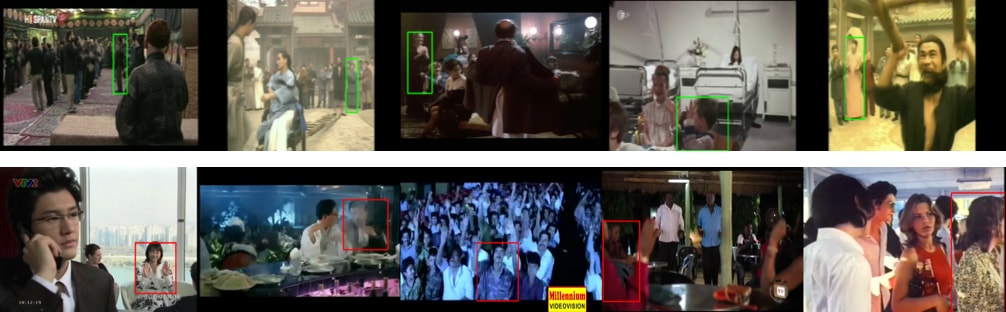}
\\
hand clap
\\
\noindent\rule{.85\textwidth}{1pt}\\[1em]
\includegraphics[width=0.8\linewidth]{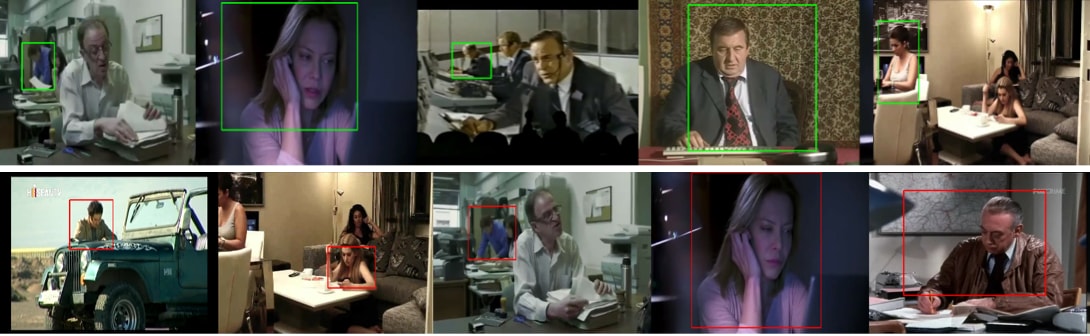}
\\
work on a computer
\\
\caption{Most confident action detections on AVA. True positives are in green, false alarms in red.}
\label{fig:top_det}
\end{figure*}

\begin{figure*}[t]
\centering
\includegraphics[width=0.8\linewidth]{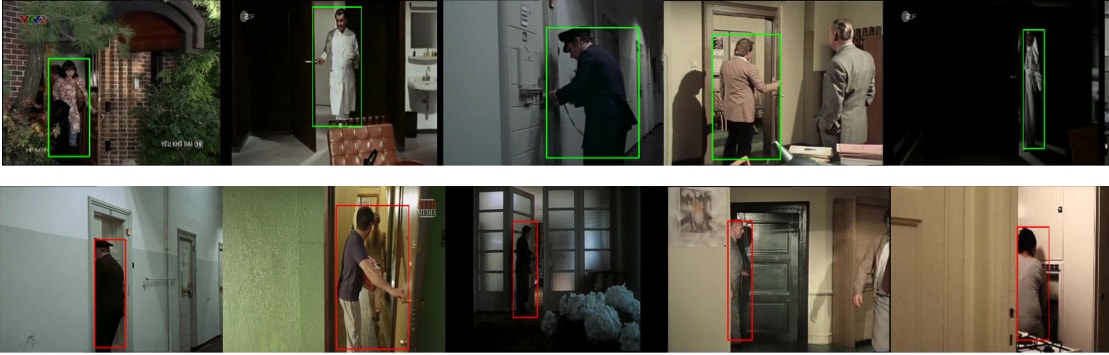}
\\
open (e.g. window, door)
\\
\noindent\rule{.85\textwidth}{1pt}\\[1em]
\includegraphics[width=0.8\linewidth]{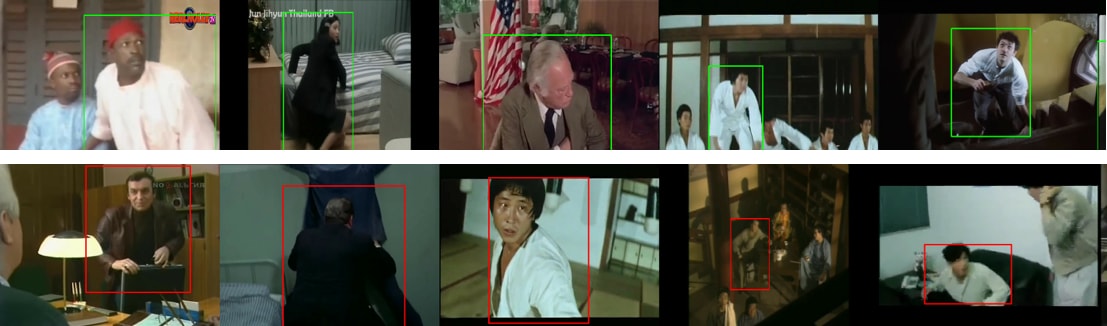}
\\
get up
\\
\noindent\rule{.85\textwidth}{1pt}\\[1em]
\includegraphics[width=0.8\linewidth]{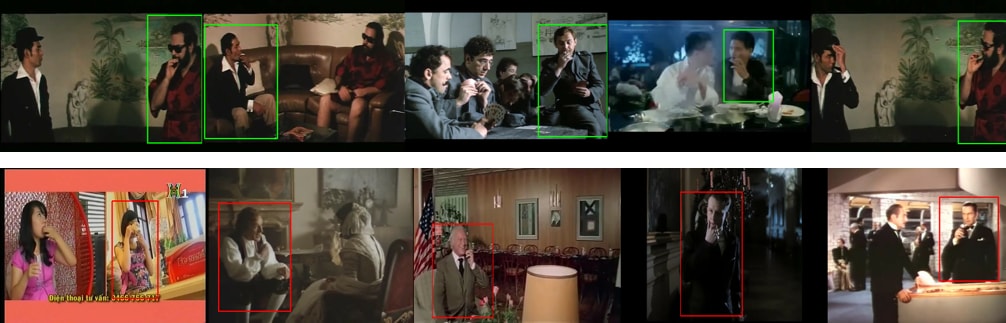}
\\
smoke
\\
\noindent\rule{.85\textwidth}{1pt}\\[1em]
\includegraphics[width=0.8\linewidth]{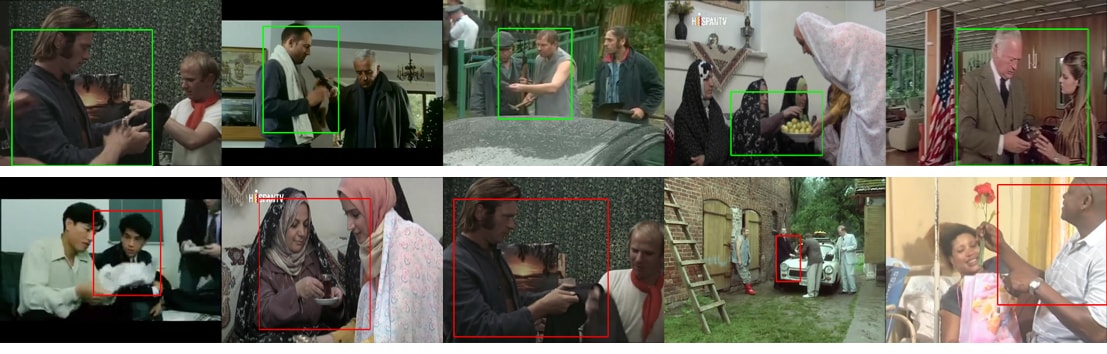}
\\
take (something) from (someone)
\\
\caption{Most confident action detections on AVA. True positives are in green, false alarms in red.}
\label{fig:top_det2}
\end{figure*}

Figure~\ref{fig:top_det} and Figure~\ref{fig:top_det2} show the top true positives and false alarms returned by our best Faster-RCNN with I3D model.

%Figure~\ref{fig:det_examples} visualizes action detection results by the same model. We show examples with high confidence, i.e., above a  threshold of $0.8$. In case of multiple labels for a human, we display the average box, if IoU  $> 0.7$.

%\begin{figure*}[h]
%\centering
%\includegraphics[width=0.8\linewidth]{supplementary_figs/example_detections.jpg}
%\caption{Example results of our action detection approach with fusion of RGB and flow with a confidence score above 0.8. In case of multiple labels of a human, we display the average box, if IoU  $> 0.7$. Different actors are identified with different colors.}
%\label{fig:det_examples}
%\end{figure*}